\documentclass{article}
\usepackage{spconf,amsmath,graphicx,hyperref}
\usepackage{booktabs}

\title{Visual Anomaly Synthesis for Model Selection under Data Scarcity}

\name{Daniel Pr\"oll$^{1,3}$, Thomas Kraxner$^{2}$, Tobias Sch\"afer$^{2}$, Sebastian Hegenbart$^{1,4}$}
\address{$^{1}$Digital Factory Vorarlberg GmbH, Dornbirn, Austria\\
         $^{2}$illwerke vkw AG, Bregenz, Austria \\
         $^{3}$University of Salzburg, Salzburg, Austria \\
         $^{4}$Vorarlberg University of Applied Sciences, Dornbirn, Austria}

\begin{document}
%
\maketitle
\begin{abstract}
In industrial use, the sensitivity of defect detection systems requires validation, yet defective samples are rare and often non-existent for a given setup. We propose a model-agnostic framework that synthesizes severity-graded defects on real defect-free images to support model selection and validation without defect references. An instruction-based image editing model generates candidate defects from literature-derived prompts; the defects are blended back into the original image, preserving unedited regions, and filtered to remove off-target edits. In a few-shot MVTec~AD study, selecting PatchCore-based anomaly detection configurations using only synthetic images, our approach nearly halves the image AUROC regret relative to a per-category oracle, compared with the best single model. A case study on monitoring Pelton turbine runners shows strong detection performance on synthetic images (AUROC 0.980), while low-severity defects still challenge the detector. The experiments highlight the need for difficult-to-detect low-severity anomalies for model selection in the absence of real defect data.

\end{abstract}
\section{Introduction}
Productive use of visual anomaly detection (AD) systems for quality control in industry is often limited by a lack of available defect data. When a new product line is first produced or first images are captured, defect-free images are easy to collect and quickly available. Modern defect detection systems are often built around unsupervised anomaly detection models, which need only nominal data for training. Before an automated system can be used in production, however, it must be validated and its thresholds selected, both of which require images of defective samples that match the specific product and image capture. The lack of defect data has motivated the use of generative models to create synthetic defect images (anomaly synthesis, AS).

We propose a reference-free image editing and model evaluation pipeline specialized for model and parameter selection without images of defective samples. The proposed model selection method is agnostic to the generative model and can be used to validate AD performance in situations where no real defect data exists, as long as the generated defects are realistic enough to challenge the detector.

The proposed method is first evaluated on the MVTec~AD benchmark~\cite{bergmannMVTecAnomalyDetection2021} and then used to develop a continuous visual defect monitoring system for Pelton turbines in hydropower, where no real defect images matching the image capture setup are available and manually creating defect images would be prohibitively expensive and likely visually unrealistic. Pelton wheel buckets wear continuously through sediment abrasion, cavitation and impact. A Pelton wheel turbine is monitored with a custom camera system (see~\cite{lotscherUseComputerVision}). To detect defects as they occur, an automated AD system is developed using only real defect-free images.

In summary, our contributions are: i) \textbf{Severity-graded, reference-free anomaly synthesis}: Defects are generated at four severity levels from a taxonomy distilled from literature, without any defective reference image or manually written prompt for the specific part. ii) \textbf{Locality-preserving edit recovery}: A feature-space change map restores the unedited regions of the source image after editing so that evaluation is not distorted by global re-synthesis. iii) \textbf{Hard-to-detect anomalies as a model selection criterion}: We show that low-severity, near-threshold defects are the most informative for ranking detectors.

\begin{figure*}[tb]
\centering
\includegraphics[width=\textwidth]{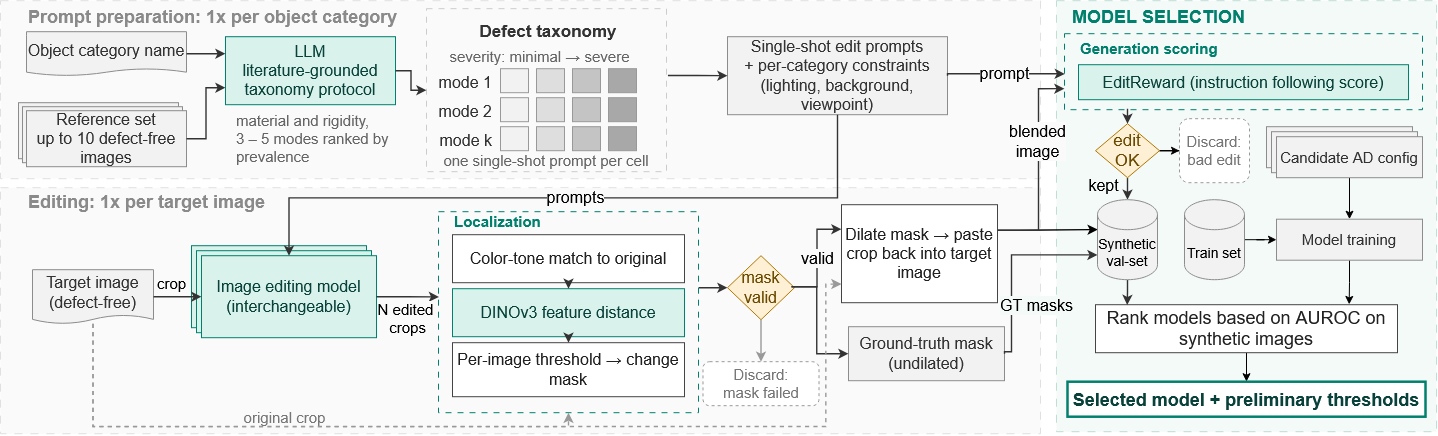}
\caption{Overview of the three steps of the proposed method: prompt and taxonomy creation (once per object), image editing and post-processing (once per image) and downstream use in evaluation and model selection. Pre-trained models shown in green.}
\label{fig:architecture}
\end{figure*}
\section{Related Work}
\textbf{Anomaly synthesis.} Anomaly synthesis methods~\cite{wangSurveyIndustrialAnomaly} differ in data requirements, realism and locality. Hand-crafted transforms~\cite{zavrtanikDRAEMDiscriminativelyTrained2021} need no defect references and apply localized edits, but are often too simple to challenge detectors. Few-shot generative methods~\cite{huAnomalyDiffusionFewshotAnomaly2024} require real defect references. Instruction-based editing models allow reference-free generation of realistic anomalies~\cite{sunUnseenVisualAnomaly2025} but usually either re-synthesize the whole image without recovering unedited regions or discard drifted outputs~\cite{kangAnomalyDetectionEffectively2026}. MIRAGE~\cite{huMIRAGEModelagnosticIndustrial2026} also derives prompts with a vision-language model and filters generated images based on edit quality, but does not grade severity or restore unedited content. Severity is usually controlled via mask size~\cite{zavrtanikDRAEMDiscriminativelyTrained2021} or a diffusion parameter~\cite{sunUnseenVisualAnomaly2025}, and wider severity coverage improves downstream performance~\cite{aramkulMultimodalControllableDefect2026} because severe edits alone are too easy to detect.

\textbf{Model selection.} Most anomaly detection benchmarks (such as MVTec~AD~\cite{bergmannMVTecAnomalyDetection2021}) assume a labeled validation set exists, which is often not the case in practice. Memory-bank-based approaches such as PatchCore~\cite{rothTotalRecallIndustrial2022} allow training-free anomaly detection. While many models can detect and localize anomalies accurately, their performance depends not only on the model architecture but also on model-specific configurations such as the backbone used for the embeddings and the layers features are extracted from~\cite{hecklerExploringImportancePretrained2023a}. To our knowledge, only SWSA~\cite{fungModelSelectionAnomaly2023} evaluates AS for model selection, reporting low selection accuracy on industrial AD datasets. We instead combine severity-graded, reference-free localized edits with difficulty-based filtering for model selection.

\section{Method}
\textbf{Prompt Preparation.} The image editing prompts are generated by a large language model (LLM; Qwen/Qwen3.8-27B~\cite{qwen2026qwen3827b}) with a fixed protocol (see the prompt preparation step in \autoref{fig:architecture}; the full protocol is available in the project repository\footnote{\url{https://github.com/dajopr/synevad}}). The LLM is prompted to derive a defect taxonomy of common defect modes from an online literature search, given only the object name and up to ten defect-free images. To create the taxonomy, the LLM first conducts an online search to classify the material, structure and rigidity of the object and to identify common defect modes. Search queries and sources are documented. The LLM then ranks the defect modes by prevalence of physical cause and visual distinctness and selects three to five distinct modes. Next, it derives a taxonomy for each selected mode, including visual characteristics, anti-patterns and common locations; the protocol requires at least two independent sources. These taxonomies are created for the two (of four) most severe defect levels. As the literature rarely characterizes minimal or early-stage signs of damage, taxonomies for the two lower severity levels are derived by softening the appearance in these descriptions. The LLM is then tasked with writing image editing prompts to add these defects to defect-free images. The prompts describe the defects at each severity level and are created for all defect modes from the taxonomy using a template containing both prescriptive and prohibitive clauses. For each object category, a constraints section to preserve lighting, background, viewpoint and image character is extracted from the defect-free samples and appended verbatim to all prompts.

\textbf{Image Editing.} The synthetic defect images are created by editing defect-free images with an instruction-based image editing model (FLUX.2 [klein]). A three-step pipeline (see the editing step in \autoref{fig:architecture}) is proposed to ensure that only relevant changes are made and the rest of the image is left unchanged. First, the overall color tone of the generated image is matched to that of the original image with a low-frequency correction. Second, because instruction-based editors such as FLUX.2 change pixels globally, pixel-wise distances cannot be used to estimate differences between the original image and the edit; instead, last-layer DINOv3 ViT-S~\cite{simeoniDINOv32025} features are computed for both images to estimate a change map. Third, an adaptive threshold on the cosine distance of L2-normalized patch tokens is calculated per image to estimate where meaningful content changes have been made. The threshold $\tau=\mu_{\mathrm{bg}}+k\sigma_{\mathrm{bg}}$ is defined as the mean cosine distance between background features plus $k=4$ standard deviations of background differences, floored at a robust noise estimate. If no mask can be created because the images are too similar, the edit is discarded. The mask is then dilated and feathered, and the masked regions of the original image are replaced with the corresponding regions of the edited image. The undilated original mask is used as the ground truth for evaluation on the synthetic data. To prevent edited patches from appearing cleaner than their surroundings, a grain-matching step estimates the sensor noise level~\cite{immerkaerFastNoiseVariance1996} and adds matching simulated noise back to the image. Because diffusion models often produce unusable outputs, images are over-generated and then sampled: image generation continues until the specified number of valid images is reached. We use the EditReward~\cite{wuEditRewardHumanAlignedReward2026} score to discard edits that did not follow instructions or did not add a defect matching the instruction to the image. The final edit score is the difference between the EditReward scores of the edited image and the unedited source image, which controls for prompt- and image-specific differences.

\textbf{Model Selection.} The synthetic anomaly dataset consists of the edited defect images, the corresponding ground-truth masks, unedited defect-free images and additional metadata for the edited images (such as edit score, defect mode and severity). This dataset is used for model selection (see the model selection step in \autoref{fig:architecture}). Multiple anomaly detection models with different configurations are fit on a set of defect-free images, and their predictions on the synthetic validation dataset are evaluated. The selected model configuration is the one with the highest area under the receiver operating characteristic curve (AUROC) on a subset of the synthetic validation dataset (such as only images with minimal severity).

\section{Results}
\begin{figure}[tb]
\centering
\includegraphics[width=\columnwidth]{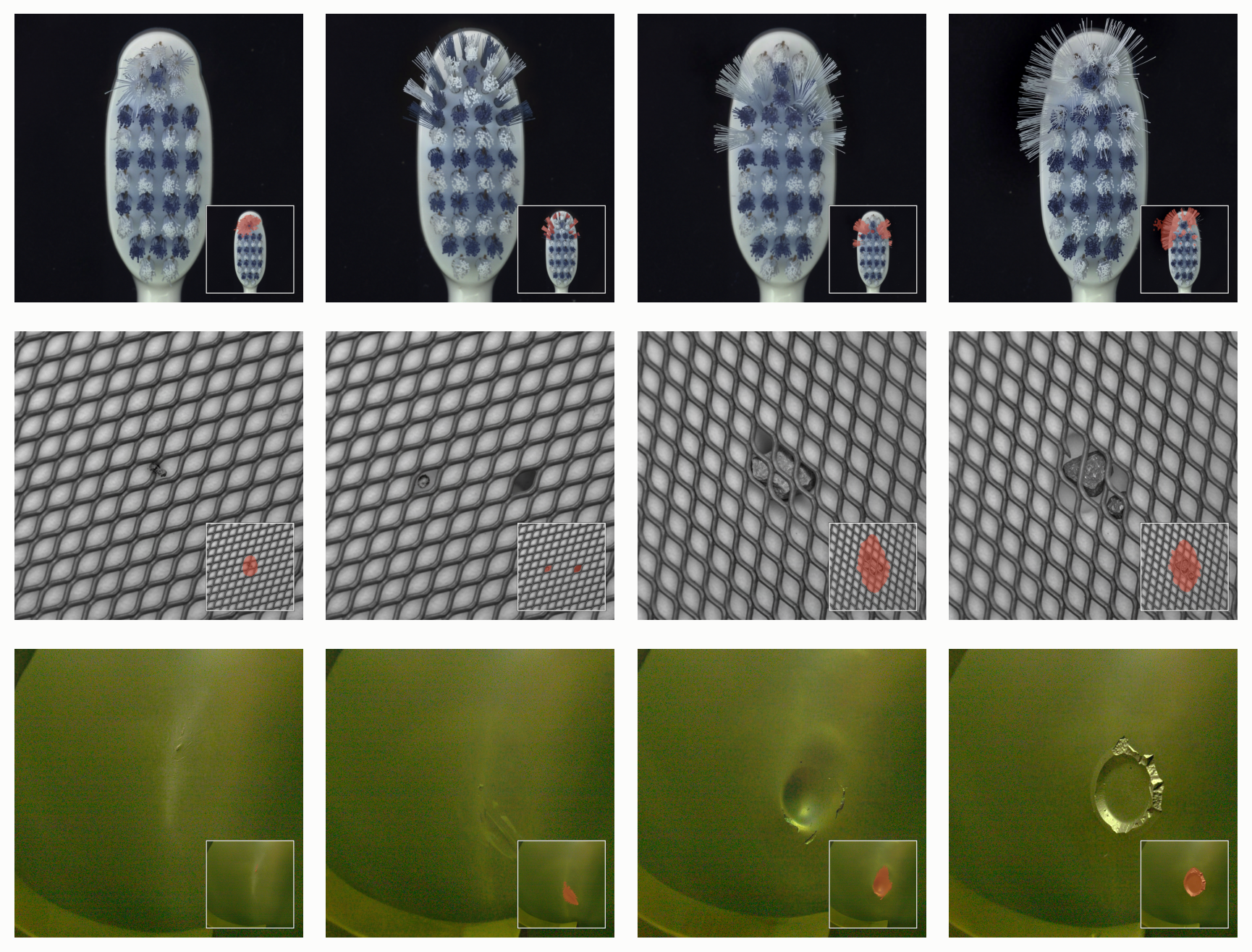}
\caption{Examples of generated images with increasing severity for two MVTec~AD categories and the Pelton use-case.}
\label{fig:gallery}
\end{figure}

\textbf{MVTec~AD.} To evaluate the predictive quality of the proposed method, we investigate few-shot MVTec~AD detection performance with 1, 2, 4 and 8 training samples over three seeds. For each object category, 20 images per defect mode and severity are generated (with a discard rate between 10\% and 20\%). \autoref{fig:gallery} shows examples of generated images at different severities with their defect masks. For model selection, different PatchCore-based configurations are evaluated on their ability to detect and localize anomalies. The configurations span reasonable model choices with pre-trained ResNet-18~\cite{he2016deep}, WideResNet-50-2~\cite{BMVC201687}, DINOv3-S and DINOv3-L as backbones, extraction from different layers (layer2+layer3 or layer2+layer3+layer4 for the ResNet models and the last layer or the last two layers for the DINOv3 models), image resolutions (256, 384), embedding dimensions (default or projection to 75\% of the default) and coreset sampling ratios (0.1 and 0.2). The model selection evaluation then compares the selected model against all models fit on the same set of data (same number of samples and seed). To contextualize prediction performance, we benchmark the model selection against two oracle baselines with access to the real test data. The first oracle selects the optimal model for each category, establishing an ideal upper bound for category-specific model selection. The second oracle identifies the single model configuration that maximizes aggregate performance across all categories; comparing against it shows how much selecting models based on synthetic data improves upon the best one-size-fits-all model. Choice regret, the difference $f(c^*)-f(\hat{c})$ in real-data performance between the oracle-selected configuration $c^*$ and the configuration $\hat{c}$ chosen by a given selection rule, is used to evaluate model selection performance.

The left side of \autoref{tab:levels-generators} shows the AD (I-AUC: image AUROC) and localization (PRO: per-region overlap between predicted and ground-truth masks; P-AUC: pixel-level AUROC) performance of the best model selected with synthetic data against the oracle baselines. The experiments show that, in the few-shot setting, model selection using synthetic data with image-level metrics improves on the best single model configuration across all categories baseline in image AUROC while achieving similar pixel-level AUROC. This improvement does not transfer to PRO, likely because edited regions are larger and more contiguous than real defects, so different configurations achieve the best PRO. Selecting models for anomaly localization requires more precise masks.
Ablations with different AS methods highlight the need for reasonably realistic AS mechanisms for model selection (right side of \autoref{tab:levels-generators}). Over a reduced configuration space (varying only backbones and coreset sampling ratios) with 60 synthetic defect images, model selection  with images generated by MIRAGE (MIR) improves AD over the best fixed model similarly to the proposed pipeline. This shows that the proposed model selection method does not depend on the specific AS model. Using the AS mechanism of DRAEM~\cite{zavrtanikDRAEMDiscriminativelyTrained2021} outperforms the fixed model substantially less, which shows the need for more realistic defects.
The left side of \autoref{tab:pelton-queries} shows the spread of the average per-category regret for models selected on different subsets of the synthetic validation data. Image AUROC regret can be nearly halved when only the lowest-severity images are used. When only moderate defects are used, performance gain is decreased. Especially the performance on the worst category is markedly improved with selecting on minimal defects (0.043 vs. 0.105).

\begin{table}[t]
\centering
\caption{Left: real anomaly detection scores of the model each rule selected, by training-set size $n$, for the best single model (Fix), from synthetic data (Ours) and by the oracle (O). Mean over categories and seeds; higher is better. Right: image AUROC choice regret, on a reduced model set, of the model selected with each generator's synthetic data, for the best single model (Fix), our pipeline (Ours), DRAEM (DRA) and MIRAGE (MIR). Mean over shots and seeds; lower is better. Bold = best within each side; oracle not ranked.}
\label{tab:levels-generators}
\small
\setlength{\tabcolsep}{2pt}
\begin{tabular}[t]{@{}lccc@{}}
\toprule
I-AUC & Fix & Ours & O \\
\midrule
1 & {0.918} & \textbf{0.927} & 0.952 \\
2 & {0.926} & \textbf{0.945} & 0.961 \\
4 & {0.938} & \textbf{0.951} & 0.965 \\
8 & {0.956} & \textbf{0.969} & 0.977 \\
mean & {0.934} & \textbf{0.948} & 0.964 \\
\midrule
PRO & Fix & Ours & O \\
1 & \textbf{0.848} & {0.842} & 0.915 \\
2 & \textbf{0.857} & {0.856} & 0.931 \\
4 & \textbf{0.894} & {0.871} & 0.943 \\
8 & \textbf{0.910} & {0.894} & 0.950 \\
mean & \textbf{0.877} & {0.866} & 0.935 \\
\midrule
P-AUC & Fix & Ours & O \\
1 & {0.938} & \textbf{0.941} & 0.977 \\
2 & \textbf{0.944} & {0.943} & 0.981 \\
4 & \textbf{0.959} & {0.954} & 0.984 \\
8 & \textbf{0.967} & {0.960} & 0.985 \\
mean & \textbf{0.952} & {0.950} & 0.982 \\
\bottomrule
\end{tabular}%
\nobreak\hspace{1em}%
\begin{tabular}[t]{@{}lcccc@{}}
\toprule
& \multicolumn{4}{c}{Image generator} \\
\cmidrule(lr){2-5}
category & Fix & Ours & DRA & MIR \\
\midrule
bottle & 0.004 & \textbf{0.001} & 0.005 & 0.003 \\
cable & 0.086 & 0.017 & 0.058 & \textbf{0.010}  \\
capsule & 0.033 & 0.033 & 0.030 & \textbf{0.011} \\
carpet & 0.003 & \textbf{0.001} & \textbf{0.001} & 0.002 \\
grid & 0.036 & 0.002 & \textbf{0.000} & 0.013 \\
hazelnut & 0.078 & \textbf{0.001} & 0.066 & 0.006 \\
leather & \textbf{0.000} & \textbf{0.000} & \textbf{0.000} & \textbf{0.000} \\
metal\_nut & 0.019 & \textbf{0.006} & 0.030 & 0.026 \\
pill & 0.039 & \textbf{0.000} & 0.081 & 0.029 \\
screw & 0.036 & \textbf{0.014} & 0.036 & 0.021 \\
tile & 0.001 & \textbf{0.000} & 0.001 & 0.002 \\
toothbrush & 0.057 & 0.032 & 0.045 & \textbf{0.011} \\
transistor & 0.084 & \textbf{0.013} & 0.022 & 0.017 \\
wood & 0.051 & 0.021 & 0.005 & \textbf{0.001} \\
zipper & 0.038 & \textbf{0.008} & 0.026 & 0.015 \\
\midrule
mean & 0.038 & \textbf{0.010} & 0.027 & 0.011\\
\bottomrule
\end{tabular}
\end{table}

\textbf{Pelton Bucket.} Since the MVTec~AD experiments show that the proposed method can be used for model selection, it is applied in a Pelton bucket case study, where no defect images matching the setup are available. The available Pelton bucket dataset consists of 190 in situ defect-free images of one turbine. For each combination of four defect modes and four severities, 20 defective images are generated, together with 320 synthetic defect-free images. PatchCore-based models with four backbones are fit on 30 defect-free training images and evaluated with the synthetic dataset, consisting of real holdout defect-free images, synthetic defect-free images and synthetic defect images. The AD models can reliably detect moderate and severe defects but are stressed by low-severity defects (see \autoref{tab:pelton-queries}). The AUROC of defect free edits (0.583) shows the models not only detect image generation artifacts. Model selection cannot be evaluated because no real defect images are available. However, the synthetic dataset enables initial use under supervision.

\begin{table}[t]
\centering
\caption{Left: MVTec~AD I-AUC choice regret for model selection based on different subsets, for the best fixed model (Fix) and synthetic anomalies of different severities from least (1) to most severe (4) and all severities (all). Median, mean and max over the per-category means (over shots and seeds); lower is better, bold = best in each column. Right: I-AUC of the synthetic Pelton defects of each mode (abr.: abrasive erosion, cav.: cavitation, crack: fatigue crack, imp.: stone impact) and severity, with synthetic negatives (defects vs.\ real crops + synthetic defect-free images), mean over four models, higher is better. Below: AUROC of synthetic negatives against real negatives. Near 0.5 is best.}
\label{tab:pelton-queries}
\small
\setlength{\tabcolsep}{2pt}
\begin{tabular}[t]{@{}lrrr@{}}
\toprule
Set & med. & mean & max \\
\midrule
Fix & 0.026 & 0.029 & 0.105 \\
1 & \textbf{0.014} & \textbf{0.016} & \textbf{0.043} \\
2 & 0.019 & 0.025 & 0.084 \\
3 & 0.020 & 0.024 & 0.069 \\
4 & {0.018} & 0.025 & 0.095 \\
All & 0.019 & {0.017} & {0.050} \\
\bottomrule
\end{tabular}%
\nobreak\hspace{1em}%
\begin{tabular}[t]{@{}lcccc|c@{}}
\toprule
Pelton & 1 & 2 & 3 & 4 & all \\
\midrule
abr. & 0.981 & 0.969 & 0.993 & 1.000 & 0.986 \\
cav. & 0.867 & 0.970 & 0.991 & 0.995 & 0.956 \\
crack & 0.969 & 0.986 & 0.992 & 0.999 & 0.986 \\
imp. & 0.975 & 0.989 & 0.998 & 1.000 & 0.990 \\
all & 0.948 & 0.978 & 0.994 & 0.998 & 0.980 \\
\noalign{\vspace{-\aboverulesep}}
\midrule 
\noalign{\vspace{-\belowrulesep}}
\multicolumn{4}{l}{defect free generations} & \multicolumn{1}{c}{0.583} \\
\bottomrule
\end{tabular}

\end{table}

\section{Conclusion}
We presented an anomaly synthesis and evaluation framework that generates severity-graded defects without defect references for a specific part and without manually written prompts, enabling model selection and validation before any real defect has been observed. Selecting a model per category on synthetic data outperforms selecting a single best configuration for all categories. Using only low-severity anomalies, image AUROC choice regret falls from 0.029 for the best fixed model---itself chosen with access to real test data---to 0.016, a reduction of 45\%. Restricting selection to moderate and severe defects recovers less of that gap (0.024 and 0.025), confirming that near-threshold defects carry most of the signal. The mechanism is not tied to our generator: over a reduced configuration space, DRAEM and MIRAGE images also improve on the corresponding fixed baseline (0.038), with MIRAGE (0.011) close to the proposed pipeline (0.010) and DRAEM improving slightly (0.027). Selection on image-level metrics does not carry over to localization, where the selected models fall behind the best fixed baseline; selecting for localization would require more precise ground-truth masks. In the Pelton bucket case study, the resulting system detects the synthetic anomalies well (AUROC 0.98) and does not merely respond to generation artifacts, while low-severity defects remain challenging. Whether it detects real defects as reliably, and whether the generated patterns match what occurs in service, can only be assessed once real in situ defect images become available. Future work will examine synthetic data for validating model training and, with improved masks, for selecting models for localization.

\vfill
\pagebreak
\section{Acknowledgments}
This work was funded by the Austrian Research Promotion Agency (FFG) under the program ``Industrienahe Dissertationen,'' grant no. FO999932323.

In writing this paper, Claude Opus 5 was used for editing and grammar enhancements.

\bibliographystyle{IEEEbib}
\bibliography{references}

\end{document}